\documentclass[acmtog]{acmart}

\usepackage{multirow}
\usepackage{amsmath}
\usepackage{bm}%%
\AtBeginDocument{%
  \providecommand\BibTeX{{%
    \normalfont B\kern-0.5em{\scshape i\kern-0.25em b}\kern-0.8em\TeX}}}

\acmSubmissionID{papers\_273}

\copyrightyear{2022}
\acmYear{2022}
\setcopyright{acmcopyright}\acmConference[SA '22 Conference Papers]{SIGGRAPH Asia 2022 Conference Papers}{December 6--9, 2022}{Daegu, Republic of Korea}
\acmBooktitle{SIGGRAPH Asia 2022 Conference Papers (SA '22 Conference Papers), December 6--9, 2022, Daegu, Republic of Korea}
\acmPrice{15.00}
\acmDOI{10.1145/3550469.3555392}
\acmISBN{978-1-4503-9470-3/22/12}

\begin{document}

\title{CLIP-Mesh: Generating textured meshes from text using pretrained image-text models}

%%Title ideas

\author{Nasir Mohammad Khalid}
\affiliation{%
  \institution{Concordia University}
  \city{Montreal}
  \country{Canada}
}
\affiliation{%
  \institution{Mila}
  \city{Montreal}
  \country{Canada}
}

\author{Tianhao Xie}
\affiliation{%
  \institution{Concordia University}
  \city{Montreal}
  \country{Canada}
}

\author{Eugene Belilovsky}
\affiliation{%
  \institution{Concordia University}
  \city{Montreal}
  \country{Canada}
}
\affiliation{%
  \institution{Mila}
  \city{Montreal}
  \country{Canada}
}

\author{Tiberiu Popa}
\affiliation{%
  \institution{Concordia University}
  \city{Montreal}
  \country{Canada}
}

\begin{abstract}
  We present a technique for zero-shot generation of a 3D model using only a target text prompt. Without any 3D supervision our method deforms the control shape of a limit subdivided surface along with its texture map and normal map to obtain a 3D asset that corresponds to the input text prompt and can be easily deployed into games or modeling applications. We rely only on a pre-trained CLIP model that compares the input text prompt with differentiably rendered images of our 3D model. While previous works have focused on stylization or required training of generative models we perform optimization on mesh parameters directly to generate shape, texture or both. To constrain the optimization to produce plausible meshes and textures we introduce a number of techniques using image augmentations and the use of a pretrained prior that generates CLIP image embeddings given a text embedding.
\end{abstract}




\begin{teaserfigure}
    \includegraphics[width=\textwidth]{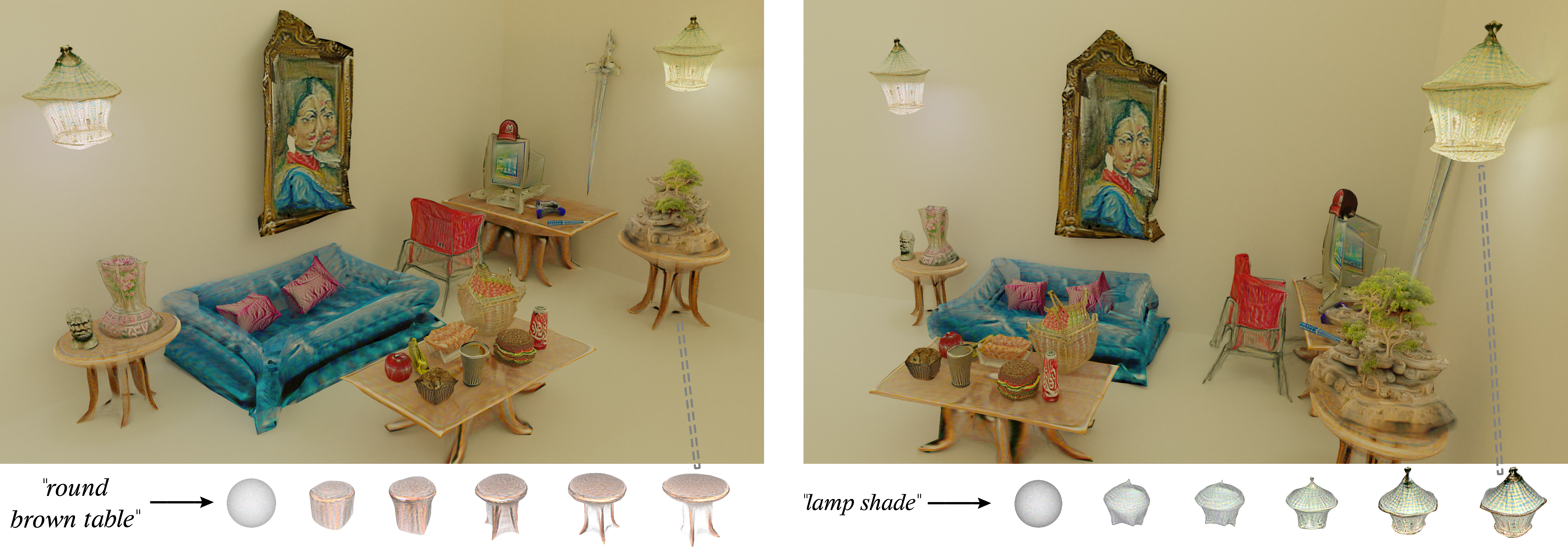}
    \caption{A 3D scene composed of objects generated using only text prompts: \emph{lamp shade, round brown table, photograph of a bust of homer, vase with pink flowers, blue sofa, pink pillow, painting in a frame, brown table, apple, banana, muffin, loaf of bread, coffee, burger, fruit basket, coca cola can, red chair, computer monitor, photo of marios cap, playstation one controller, blue pen, excalibur sword, matte painting of a bonsai tree; trending on artstation}. (The 3D positioning in the scene was done by a user)}
    \Description{A 3D scene composed of objects generated using only text prompts: \emph{lamp shade, round brown table, photograph of a bust of homer, vase with pink flowers, blue sofa, pink pillow, painting in a frame, brown table, apple, banana, muffin, loaf of bread, coffee, burger, fruit basket, coca cola can, red chair, computer monitor, photo of marios cap, playstation one controller, blue pen, excalibur sword, matte painting of a bonsai tree; trending on artstation}. (The 3D positioning in the scene was done by a user)}
    \label{fig:teaser}
\end{teaserfigure}

\maketitle

\section{Introduction}

%- Mention some applications - e.g. gaming worlds
Gaming, virtual reality, films and other multimedia experiences rely on the use of 3D models. While there are various methods of representing these models, many existing games and modeling software use 3D assets consisting of a polygonal mesh coupled with texture and normal maps. However, the creation and texturing of meshes is a time consuming and expensive task that often also needs specialized software. 
There has been a lot of research focused on synthesizing shapes but these look at generation in the form of point clouds, voxel grids or implicit functions and are restricted to fixed shape categories. 
While these provide good results the issue is they require additional steps to convert to meshes that can be used in existing software and this conversion can lead to undesirable results or artifacts.

%- Generating 3d objects from abstract text descriptions is an important open problem 

The ideal scenario would be a technique where a user can generate any arbitrary 3D shape based on only an abstract text description of the object. This would greatly increase the use and accessibility of developing 3D assets. Furthermore, if the shape generated is in the form of a mesh with corresponding texture maps it would easily facilitate integration with a large suite of existing game engines and software that use 3D meshes as primitives.

%- Availability of large datasets of 3D object examples and corresponding natural language is very limited compared to the rich 2D images and text descriptions available. 

%- This brings the question, Can we infer 3D shape and texture using the knowledge from Large scale Deep Learning models trained only on images and text

A big limitation is the lack of large varied datasets of 3D objects and corresponding natural language descriptions. For example, datasets such as Shapenet \cite{chang2015shapenet} and CO3D \cite{reizenstein21co3d} provide 50 object categories respectively. In contrast there are large datasets containing rich 2D images with a large variety of objects. For example Imagenet-21K \cite{ridnik2021imagenet21k} has 21,000 object categories. Furthermore, natural image data can often be accompanied by rich textual descriptions. Recently CLIP has been trained on a large dataset of 400 million image text pairs to learn an aligned visual and textual representation \cite{radford2021learning}. This text and image scoring model was trained on text captions with combinations from a set of 500,000 query words, leading to a very large diversity in the potential objects it can represent.

 We thus consider utilizing the knowledge from a large scale deep learning model trained only on images and texts. This relies on the fact that a 3D shape can be projected to a 2D image from an arbitrary viewpoint through rendering. Using a differentiable renderer such as \cite{nvdiff} one can obtain images of a shape and then use CLIP to get a score between the images and an input text. Leveraging the differentiability of the renderer and CLIP, an inverse problem can be solved by optimizing the shape and texture of a mesh to maximize the CLIP score of rendered images and input prompt. 
 However, doing this naively can lead to a tangled and noisy mesh as there are insufficient constraints on the shape. Therefore we incorporate a number of constraints and techniques that allow us to generate a plausible shape and texture. 
 % Tibi: question --> do we use regularization loss o top of subdivision?
 First we use a regularization loss and incorporate limit subdivision to further smooth the mesh. Even though this helps us maximize the score it often leads to an undesirable result in terms of texture as CLIP may prefer "painting" small artifacts in to the texture rather than deform and globally texture the object. To alleviate this we use multiple augmentations to render the object dynamically such that the optimizitation reaches a solution leveraging the shape information. Additionally to further improve results we introduce a conditional generative model that uses a pretrained diffusion prior model, that generates a CLIP image embedding given the text prompt, this is similar to current state of the art text to image synthesis work \cite{Ramesh2022HierarchicalTI}. Our contributions can be summarized as followed:
 \raggedbottom
\begin{itemize} %Incorporate emphasis on subdivision surface
    \item We introduce a set of techniques that allow zero-shot text-guided generation with a differentiable renderer.
    \item We use these techniques to directly generate 3D meshes with their texture maps and normal maps .
    \item We use the analytical expression of the Loop subdivision limit surface as an implicit regularizer to improve the quality of the generated model.
    \item We improve on our baseline results by introducing a set of render augmentations and incorporating a text to image embedding prior.
\end{itemize}

\section{Related work}

A variety of recent works focus on text driven 2D image manipulation and generation using CLIP \cite{radford2021learning}, a model that learns a joint embedding space for image and text. Leveraging CLIPs joint embedding, many works such as StyleCLIP \cite{Patashnik_2021_ICCV}, VQGAN-CLIP \cite{crowson2022vqgan} and GLIDE \cite{nichol2021glide} have shown that pretrained image generative models can be guided by text prompts through distance losses in the shared embedding space. Additionally, the current state of the art in text to image generation trains a model directly on CLIP text and image embeddings \cite{Ramesh2022HierarchicalTI}. 

% Discuss 3D object generation with deep learning in general from works using different representations (voxel representations, SDF). Generation using differentiable renderers 

In contrast to this, text to 3D is an underdeveloped field but a number of works have previously attempted to generate 3D models from text by utilizing datasets of text descriptions corresponding to 3D models. For example \cite{chen2018text2shape,fukamizu2019generation} proposed to train a joint embedding between 3D shapes and text and combine this with generative adversarial networks \cite{GAN} to produce novel outputs. These approaches however are not zero-shot and are thus limited by the lack of available matched 3D models and text descriptions. CLIP-Forge \cite{Sanghi2021CLIPForgeTZ} alleviates the issue of paired text and 3D models by relying only on the 3D models to train an encoder and decoder and then guiding generation of the decoder with CLIP to produce results that match a text prompt, this only partially solves the problem because now the generation is restricted by the 3D data categories available to train it. It also doesn’t produce meshes or textures and thus its use is limited. \cite{avatarclip, jetchev2021clipmatrix} focus on stylization of predefined human shape to match an input text prompt and \cite{michel2021text2mesh} generalizes this to any arbritary mesh and text prompt. Text2Mesh addresses a related but different problem: with a correct starting mesh each vertex is minimally modified along the normal direction and its color. 
%The authors of Text2Mesh themselves are very clear in their Methodology section 3.1 that their architecture is not suited for generalized shape optimization as we are attempting: “This formulation tightly couples the style field to the source mesh, enabling only slight geometric modifications”. Therefore, a direct comparison of results is not possible and a comparison of performance is not relevant.

% - Is not Zero-shot - requires training pairs - not as scalable
% - For both these works use the word "concurrent work"

Dreamfields \cite{jain2021zero} proposed a zero-shot text guided generation using a NeRF model \cite{mildenhall2020nerf}. Unlike our approach this does not allow direct generation of a mesh but instead trains a neural radiance field. This method requires raycasting and training a set of neural network parameters which has a large computation overhead even for low quality generation where as our figures are all generated on a single 16GB GPU. Additionally editing of the object and getting a mesh is not straightforward since the shape is within the weights of a network and extraction requires a user determined thresholding which can lead to trade offs. Furthermore, the texture and shape cannot be disentangled. While in our work the shape, texture and normal can be individually modified allowing unique application scenarios. For example we demonstrate multi object optimization within a scene which is straightforward under our method but the same cannot be easily applied using \cite{jain2021zero}.% can also talk about normal map and texture disconnect but feels like im nitpicking

% - Work is recent - does not provide a mesh - requires a separate parameterized NERF model

%Discuss CLIP

\begin{figure*}
\begin{center}
\includegraphics[width=0.9\linewidth]{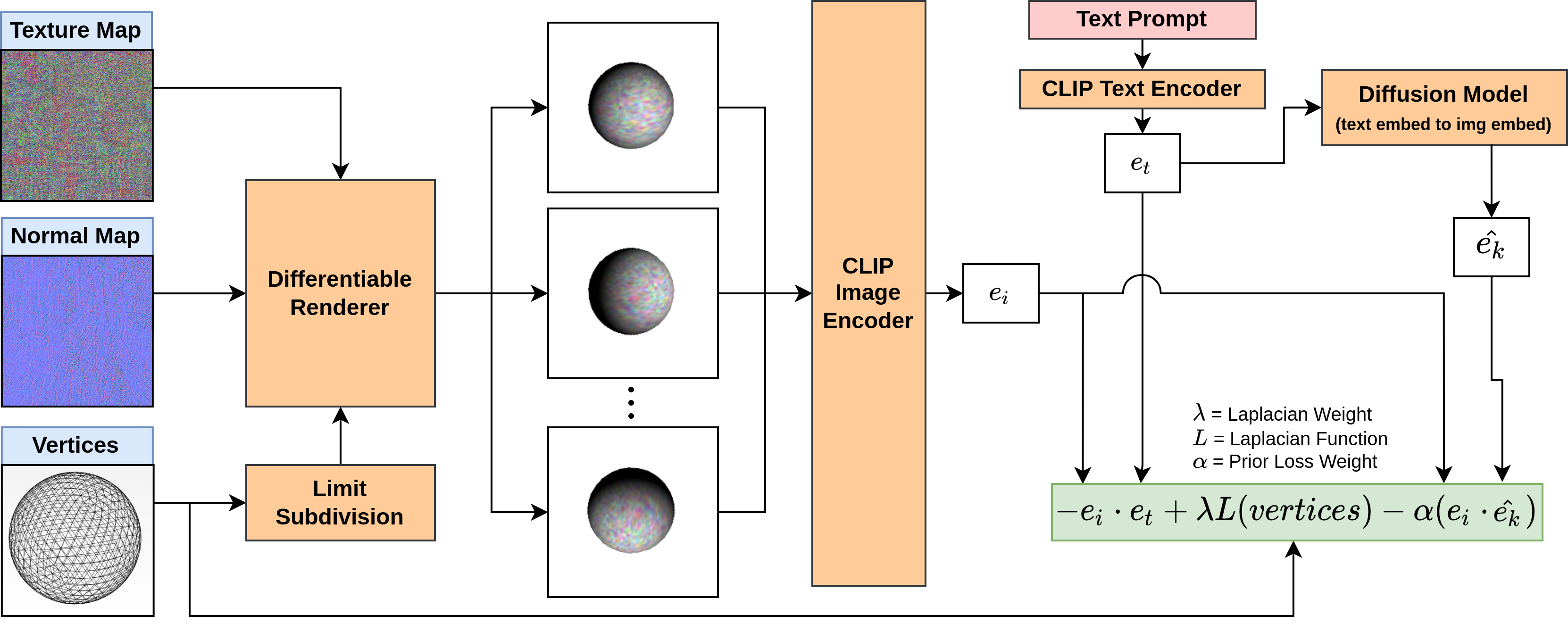}
\end{center}
   \caption{Overview of our optimization pipeline. The differentiable renderer creates views which are encoded and compared to the text encoding as well as the generated image embedding. We optimize for the texture, normal, vertices position.}
\label{fig:overview}
\end{figure*}

% \subsection{Text-Guided 3D Shape Generation with CLIP }
\vspace{-3pt}\section{Method}

%Describe the full objective function used to optimize and reference the main figure
An overview of our method is shown in Figure~\ref{fig:overview}. 
We represent a 3D model using three components: (1) a 3D mesh whose vertices $ \bm{V}_0 \in \mathbb{R}^{n \times 3} $ are the control vertices of a Loop~\cite{loop1987smooth} subdivision surface $\bm{V}=S(\bm{V}_0)$, (2) a texture map $T$ and (3) a normal map $\tilde{\bm{T}}$.
This is a standard way to represent geometric assets in video games and modeling applications.
Furthermore, using a texture map allows to decouple the appearance from the geometry and the combination of normal map and subdivision surface control allows us to reduce the number of optimization parameters of the geometry while maintaining rendering details. 
Our method creates a 3D model by optimizing these three components using a differentiable renderer. 
Our rendering pipeline uses the initial control mesh to compute the limit surface $V$ of the Loop subdivision scheme~\cite{stam1998evaluation}. This limit surface can be computed analytically and it is a differentiable function. The loop subdivision surface $V$ is also, by construction, smooth. Therefore, this surface definition acts as an implicit regularizer and helps avoid triangle inversion during the optimization phase.
We render this mesh using a differentiable renderer $R$~\cite{nvdiff} from several camera positions $D(\varphi, \theta)$.
%, which is smooth by definition.  the Loop subdivision followed by the rendering of the resulting dense 3D mesh using a color texture map $ T_t \in [0, 1] ^{H \times W \times 4} $ and a normal map $ T_n \in [-1, 1] ^{H \times W \times 3}$. 
%As the loop subdivision step is differentiable 
%primitive 3D control shape given by $C$ that has a fixed number of vertices $ V_0 \in \mathbb{R}^{n \times 3} $, faces $ F \in \{1, ..., n\}^{m \times 3} $ and a set of UV coordinates for texturing $ U \in \mathbb{R}^{n \times 2} $ which provide a lookup point for each vertex in the mesh. 
%The height and width of the maps are hyper parameters that can be tuned based on user preference. Optimization is performed on the texture maps and vertices $ V_0 $, but the vertices are first loop limit subdivided to obtain a smoother set $ V \in \mathbb{R}^{n \times 3} $ which is used in the rendering process along with the textures.
%distance $d$ along the Z axis (Section~\ref{}). 
We  uniformly sample a camera azimuth angle $\varphi$ from a range of 0$^{\circ}$  to 360$^{\circ}$ and for elevation $\theta$ we sample from a Beta distribution with  $ \alpha_d = 1.0 $ and $ \beta_d = 5.0 $  within a range of 0$^{\circ}$  to 100$^{\circ}$ this allows the generation to focus on making the object consistent from a single elevation angle giving it a "front view" but the distribution allows other elevations so that textures get painted in for triangles in those regions but the shape does not deform significantly. 
Using these camera positions and orientation we render a set of images $I$:
\[
    \bm{I} = R(D(\varphi_i, \theta_i), \bm{V}, \bm{T}, \bm{\tilde{T}})
\]
Images $I_i$ are encoded using the CLIP image encoder $C^I$:
\[
\bm{E}=C^I(\bm{I})
\]

Where $E$ represents a set of encodings for each image in $I$. The input to our method is a text prompt $\bm{p}$ that is encoded using the CLIP text encoder $C^T$:
\[
\bm{e}_t = C^T(\bm{p})
\]
As the rendered images as well as the text prompt are now encoded in the same space we can 
compute the similarity:
%$L_{CLIP}$ between $C_i$ a where $C_I$ is the CLIP image encoder which takes an input image,  $\bm{i_d} \in I_d$ and produces its normalized encoding, $\bm{e}_i =  C_I(\bm{i_d})$. Similarly, $C_T$ is the CLIP text encoder which takes a test input $\bm{s}$ and produces a normalized text encoding, $\bm{e}_t =  C_T(\bm{s})$. We can now describe our objective function as the negative of the average similarity score between the embeddings of each view and the embedding of the text prompt

\begin{equation}
    L_{CLIP}(\bm{V},\bm{T},\bm{\tilde{T}},\bm{p}) = -\frac{1}{K}\sum_{\bm{e}_i \in \bm{E}}{\bm{e}_i^T \bm{e}_t}
\end{equation}

%where $i$ iterates over the images $\bm{I}_i$. 
Note that the encoder functions, $C^T$ and $C^I$, include a normalization at the end thus these are cosine similarities. As computing the limit loop subdivision surface is differentiable~\cite{stam1998evaluation} and the renderer is differentiable, our entire pipeline is differentiable using the chain rule. 

\begin{figure*}
\begin{center}
\includegraphics[width=0.8\linewidth]{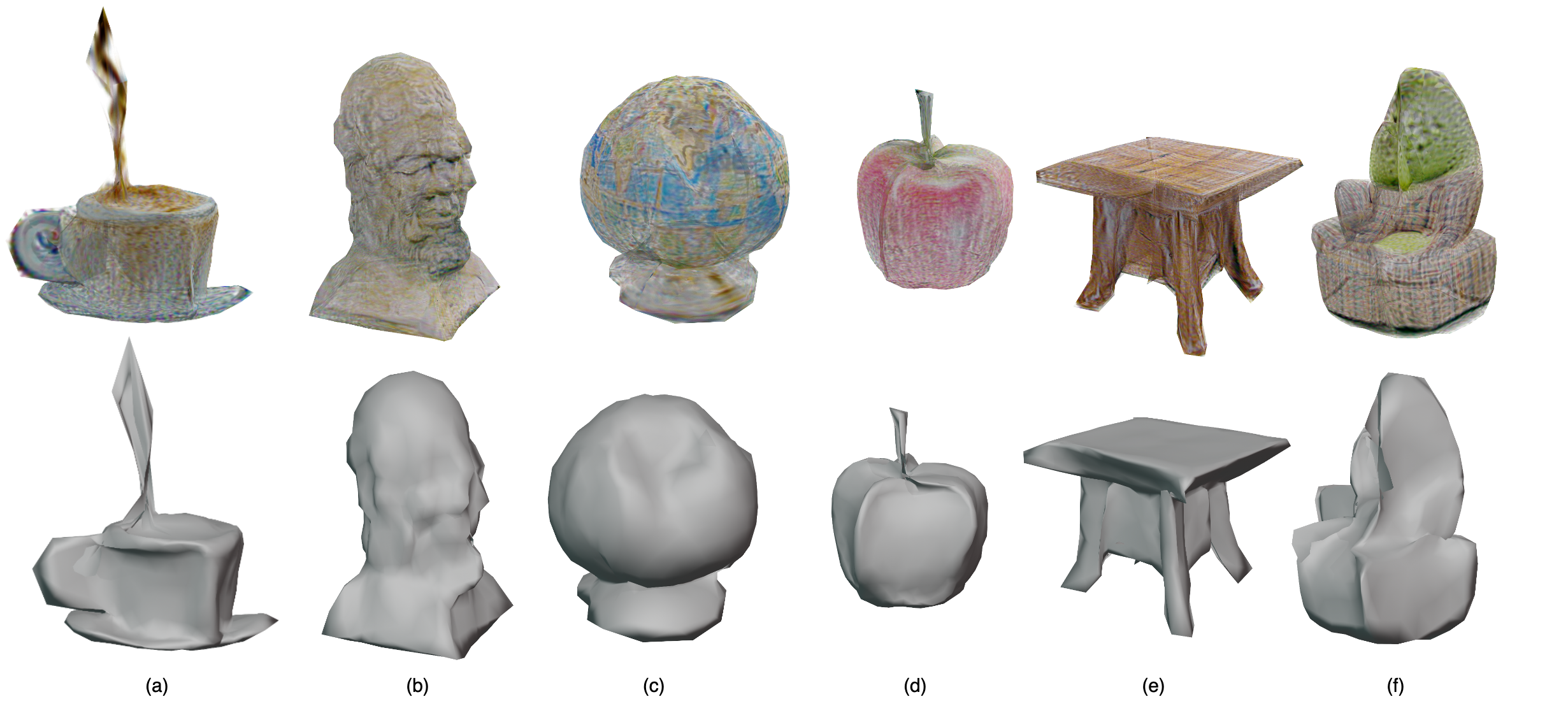}
\end{center}
   \caption{Results from a wide variety of prompts. Top: rendered result. Bottom: 3D mesh. a) "a coffee" b) "a photograph of a bust of homer" c) "Globe" d) "a apple" e) "a brown table" f) "an armchair in the shape of an avocado"}
\label{fig:result_salad}
\end{figure*}

\paragraph{Laplacian Regularizer} We use a laplacian regularizer on the shape of the mesh to maintain the geometry and keep it intact as used in other related work \cite{Hasselgren2021}. 
We use the uniformly-weighted Laplacian operator: $ \delta _i = v_i - \frac{1}{|N_i|} \sum_{j \in N_i } v_j  $  where $N_i$ is the set of one-ring neighbours for vertex $v_i$. With this formulation the laplacian regularizer can be given by:
\begin{equation}
    L_{\delta} = \frac{1}{N} \sum_{i=1}^{N} \Vert \delta _i  \Vert ^2
\end{equation}
where $N$ is the number of vertices.
This minimizes the difference in position between each vertex and the average position of its neighbouring vertices. 

\paragraph{Diffusion Prior} To further improve results we also train and incorporate a diffusion prior which attempts to generate image embeddings following $p(e_i|e_t)$.  We use this to sample image embeddings given a text encoding. Our formulation follows that of \cite{Ramesh2022HierarchicalTI} and \cite{ho2020denoising}. Once trained, the diffusion sampling process takes input of noise and the CLIP text embedding $\bm{e}_t$ and after applying the forward process for N timesteps the output is a CLIP image embedding which follows $p(e_i|e_t)$.

%$$ \bm{\hat{e}}_k = \prod_{n=1}^{N} f_\theta( \bm{\hat{e}}_{k_{(n)}} , n, \bm{e}_t )$$

We pretrain this prior on a 400 million image and text pair dataset \cite{Schuhmann2021LAION400MOD} so it can sample a relevant CLIP image embedding when given a CLIP text embedding and during optimization time we sample from it using the previously obtained text embedding $\bm{e}_t$ to get a relevant CLIP image embedding $\bm{\hat{e}_k}$. As the rendered images are encoded in the same space as the output embedding we can also compute a similarity between them to use as a loss.

\begin{equation}
    L_{PRIOR}(\bm{V},\bm{T},\bm{\tilde{T}},\bm{p}) = -\frac{1}{K}\sum_{\bm{e}_i \in E}{\bm{e}_i^T \bm{\hat{e}}_k}
\end{equation}

%This loss encourages the rendered image to a distribution of real images
Since it is conditioned on the text embeding we can use $L_{PRIOR}$  without $L_{CLIP}$. Practically, in our preliminary experiments we found that combining these losses can be beneficial. 

We thus formulate our final problem as an optimization problem with the following objective function:
\begin{equation}
     \min_{\bm{V}_0,\bm{T},\bm{\tilde{T}}} L_{CLIP}(S(\bm{V}_0),\bm{T},\bm{\tilde{T}},\bm{p}) + \lambda_t L_{\delta}(\bm{V}) + \alpha L_{PRIOR}(S(\bm{V}_0),\bm{T},\bm{\tilde{T}},\bm{p})
\end{equation}

\begin{figure*}
\begin{center}
\includegraphics[width=\linewidth]{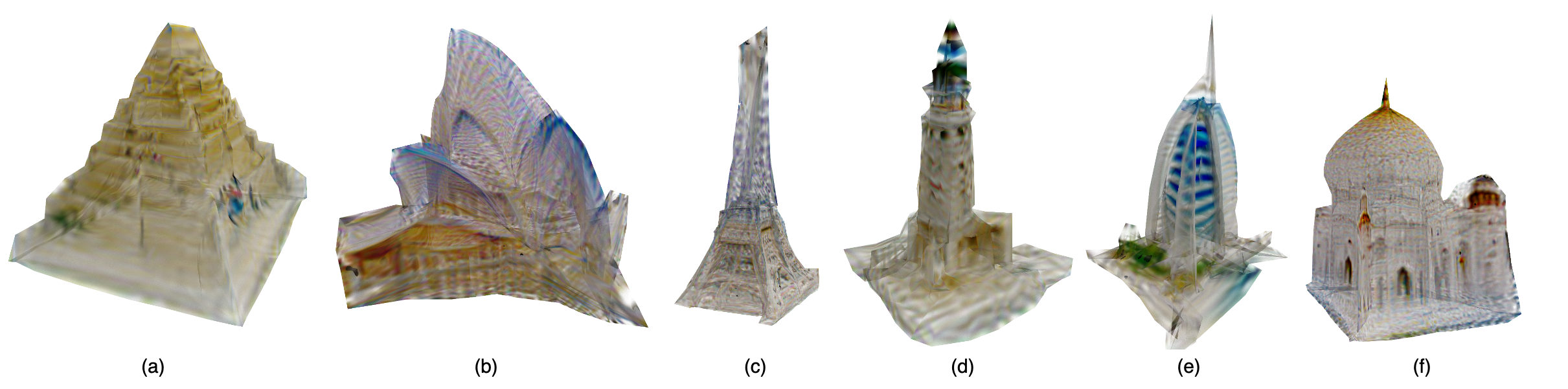}
\end{center}
   \caption{Reconstruction of famous landmark around the world: a) "pyramid of giza " b) "Sydney opera house" c) "Eiffel Tower" d) "lighthouse of alexandria" e) "Burj Al Arab" f) "Taj Mahal"}
\label{fig:result_wonders}
\end{figure*}

\paragraph{Practical Considerations and Implementation Details}

Our initial shape is a sphere with 600 vertices. The texture map is initialized with random values and is set to a resolution of 512x512. The normal map has the same resolution but is initialized as a uniform blue image. Adam optimizer is used for the vertices and texture maps with a decaying learning starting at 0.001 and a batch size of 25. The diffusion prior follows the same configuration setup as \cite{Ramesh2022HierarchicalTI} except  ours is scaled down.

The approach for the laplacian regularization follows that of \cite{Hasselgren2021}, where the weight, $\lambda$, is decayed throughout the optimization process as the shape stabilizes its final form. Initially it is set to a high value when the learning rate is high and then slowly reduces to a minimum value. More specifically, for an epoch $ t $ it is defined as $ \lambda_t = ( \lambda_{t-1} - \lambda_{min} ) \cdot 10^{-kt} + \lambda_{min} $. The initial weight and decay parameters are hyperparameters that can be tuned.

%\paragraph{Encouraging Implementation Details.}
The look-at and up vectors of the cameras are set towards the origin and the y-axis respectively. Due to the known texture bias of visual recognition models such as CLIP \cite{geirhos2018imagenettrained} naively performing the optimization can lead to over emphasis on the texture versus shape. To deal with this we add in some randomization to the view generation process by randomly selecting a camera field of view between  $30^{\circ}$ to $60^{\circ}$ and varying the distance of the camera from the object to between 3.0 to 7.0. This variance in the field of view and distance has a zoom in/out effect that encourages changes in the vertex positions versus only changes in the texture. CLIP takes 224x224 input images but we find that rendering at a larger 512x512 resolution and down scaling to 224x224 improves results, it also plays well with the differentiable render we use \cite{nvdiff} since it relies on anti aliasing for gradients and rendering at a larger resolution means more pixels are affected by anti aliasing which reduces gradient noise.

\paragraph{Random Augmentations}

To further improve results we use augmentations throughout the optimization process for Eq. 4. At each iteration we render images with a random background  similar to \cite{jain2021zero} using either gaussian noise, a solid color or a random color checkboard pattern. Additionally, we randomly offset the object so that it is not always positioned in the center of the image. These augmentations are used so that the CLIP optimization does not rely on the background or position of the object to minimize the loss but is instead forced to generate a shape that is coherent under all conditions.

\begin{figure}
\begin{center}
\includegraphics[width=1.0\linewidth]{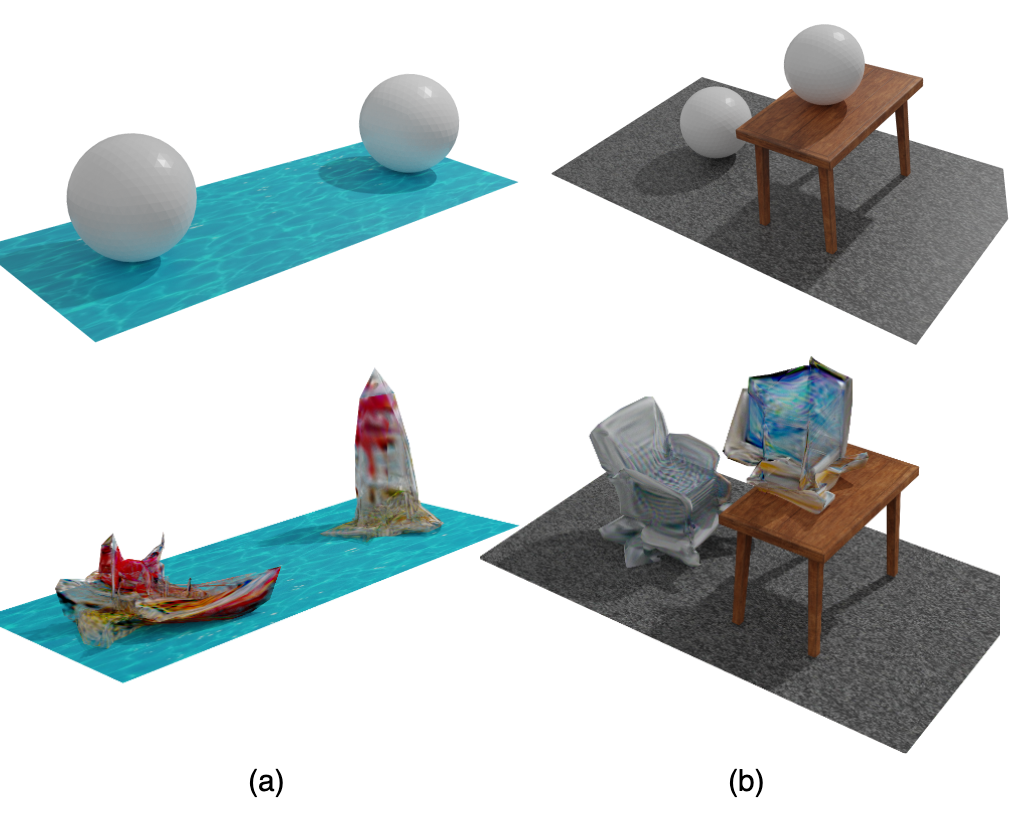}
\end{center}
   \caption{Multiple object optimization. Prompts: a) "boat and red lighthouse" b) "office chair and a desk and a computer monitor" }
\label{fig:result4b}
\end{figure}

\begin{figure*}
\begin{center}
\includegraphics[width=1.0\linewidth]{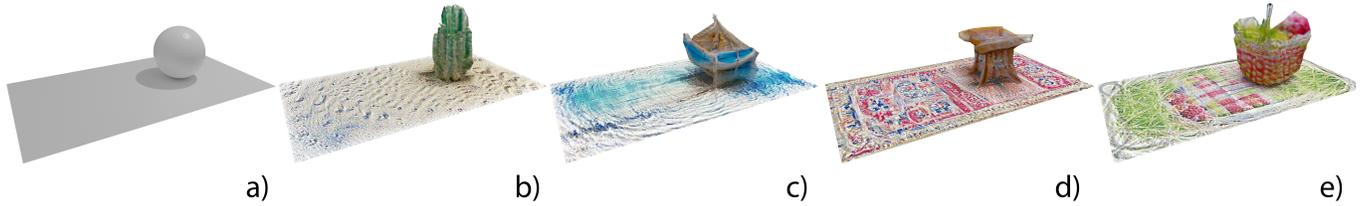}
\end{center}
   \caption{Multiple object optimization where one of the objects has fixed shape. a) initial shapes. The following are results of the following captions: b) "cactus and sand" c) "wooden boat and blue water" d) "brown wooden table and iranian carpet" e) "fruit basket on grass"}
\label{fig:result4c}
\end{figure*}

\begin{figure*}
\begin{center}
\includegraphics[width=0.8\linewidth]{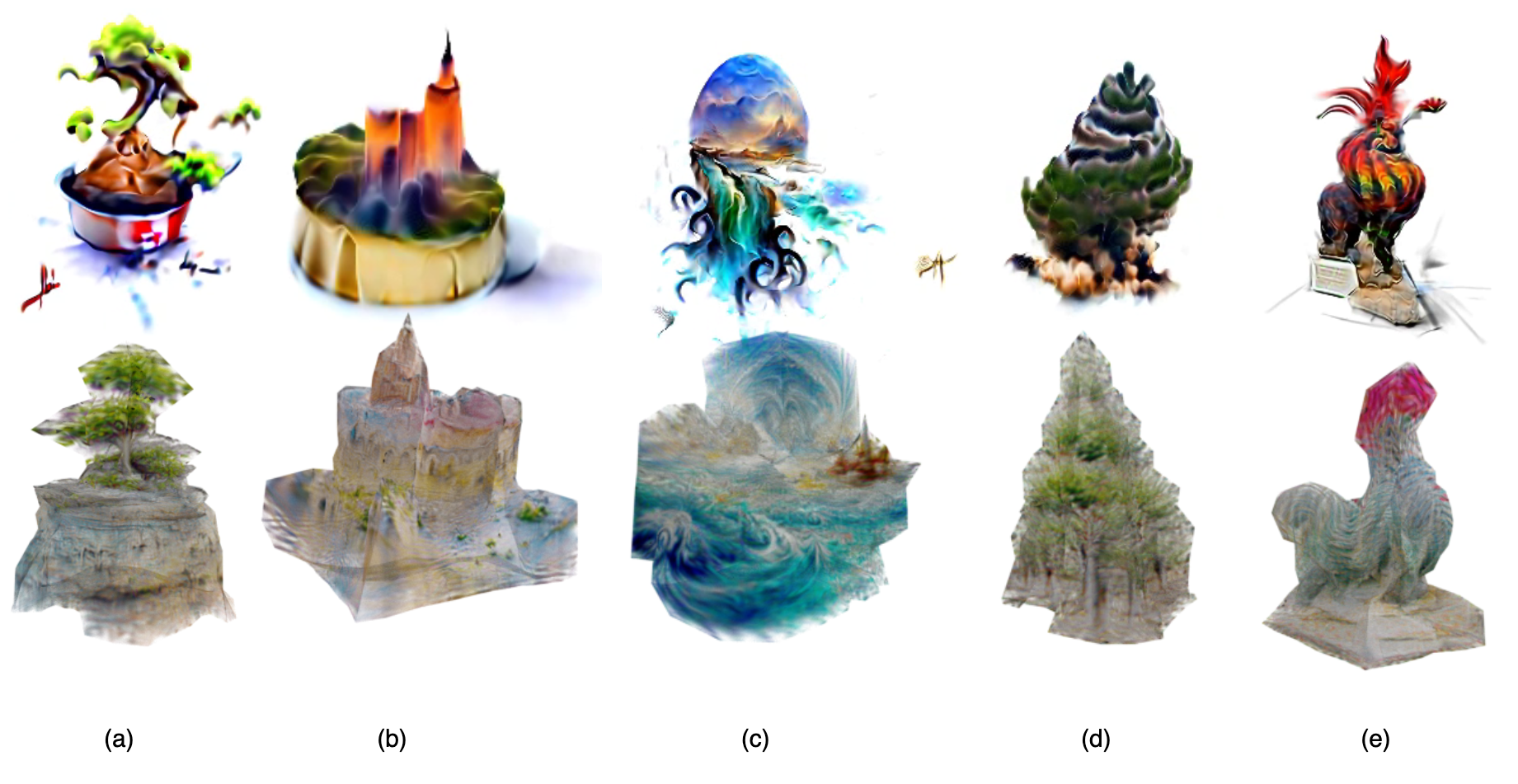}
\end{center}
   \caption{Comparison with~\cite{jain2021zero} results from their paper/project website. Top: results from ~\cite{jain2021zero}. Bottom: our results. Prompts: a) "matte painting of a bonsai
tree; trending on art station" b) "matte painting of a castle made of cheesecake surrounded by a moat made of ice cream; trending on artstation; unreal engine" c) "a cluster of pine trees
are in a barren area" d) "a cluster of pine trees
are in a barren area" e) "a sculpture of a rooster"}
\label{fig:result_comp1}
\end{figure*}

\begin{figure*}
\begin{center}
\includegraphics[width=0.8\linewidth]{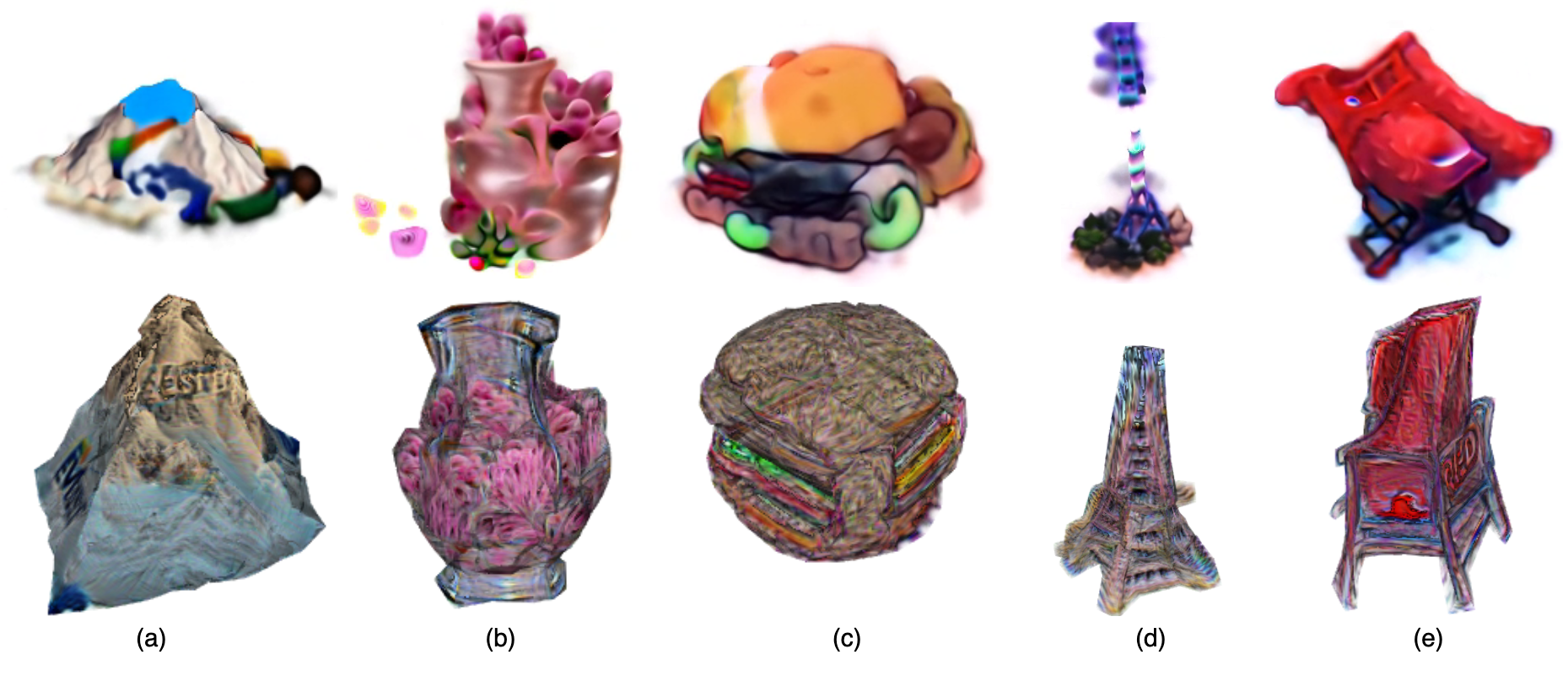}
\end{center}
   \caption{Comparison with~\cite{jain2021zero}. Shapes generated using CLIP ViT/B-16 Top: results from ~\cite{jain2021zero}. Bottom: our results. Prompts:  a) "mount everest" b) "a vase with pink flowers" c) "a hamburger" d) "Eiffel tower" e) "a red chair" }
\label{fig:result_comp2}
\end{figure*}

\section{Results and Evaluations}

%subsection 1  Discuss fig 1, fig 3 basic results, strengths weakness etc
%subsection 2 comparisons to Jain 
%subsection 3 optimization scenarios
%subsection 4 Ablations

We evaluated our methods on a wide variety of prompts and a few different generation scenarios. 
We first look at the single object generation scenario and compare our method with Jain et al.~\cite{jain2021zero}. 
We then follow up with additional modeling scenarios unique to our method.
Finally we provide quantitative evaluations of our results as well as ablation studies to illustrate the improvement provided by each step of our method.

\subsection{Single Object Generation}
In Figure~\ref{fig:teaser} we illustrate a number of household objects  generated using the proposed method. The flexibility of the assets created is illustrated as we import and place them into a 3D scene. In Figure~\ref{fig:result_salad} we further illustrate a diverse set of objects and their corresponding shape (removing the texture). Finally in Figure~\ref{fig:result_wonders} we further show the diversity of possible objects that can be generated using the knowledge of the CLIP model by producing famous landmarks  which are visually recognizable. In all these figures we use the CLIP ViT/B-32 model for training.

%\textcolor{red}{DISCUSS SHAPE ONLY OPTIMIZATION (how it is achieved and the results)}
%Discuss shape Fig 5

We also provided visual comparisons to \cite{jain2021zero}.  Fig.~\ref{fig:result_comp1} shows the results of our methods results with five prompts from \cite{jain2021zero} with the results shared in their paper and project website. We render the meshes from similar angles.
Fig.~\ref{fig:result_comp1} shows a second comparison with \cite{jain2021zero} where we chose new prompts and generated the results using the code available online. Note that because their work uses a NeRF representation and requires ray casting it comes with a large resource constraint. Therefore we use the smallest CLIP ViT-B/16 model for the generations and use the medium quality configuration provided in their codebase.

In terms of speed our method is much faster than Dreamfields \cite{jain2021zero} where each shape took over 24 hours to generate using 4 NVIDIA A100 GPUs. 
%We try to match the hyper parameters of Dreamfields as best as possible.
For similar configurations our experiments revealed that our method is faster by a factor of ~100 as each of our shapes required 50 minutes on a single NVIDIA P100 (16GB) GPU. In short the reason for this is two-fold: 1) the number of optimizing parameters in Dreamfields is much higher (all the weights of a complex neural network as opposed to vertex positions, texture and normal maps) 2) our rasterization based rendering is much faster.

\subsection{Complex Modeling Scenarios}
Another powerful feature of our method (and unique among NERF based approaches such as \cite{jain2021zero}) is the flexibility of our optimization framework. The texture and shape are decoupled allowing us to selectively optimize them if needed, and to generate multiple objects in context. This provides a number of unique possibilities for user control of the generation. Additionally since we use meshes it is trivial to combine multiple meshes in to a single mesh while also freezing some vertices and allowing others to be optimized. All this allows us to perform simultaneous optimization of multiple objects as well as separate the shape and texture optimization. This can be be very useful when modeling a scene where some objects have fixed shape while other objects are allowed to vary.

Figure~\ref{fig:result4b} shows an example of this multiple object optimization. In Figure~\ref{fig:result4b}a) the text caption used was "boat and red lighthouse", the initial setup was a plane with fixed water texture and 2 spheres on either ends. Vertices and texture for the water were frozen but spheres allowed to optimize. The final result created two distinct shapes for each object in the caption that fits the scene.
In Figure~\ref{fig:result4b}b) a similar setup is followed where the carpet and table are static, but the chair and computer monitor are automatically generated from initial spheres. Note that while the starting position of one of the sphere was on the table, we did not specify anywhere explicitly that the monitor should be on the top of the table or that the chair must face the monitor, all of this was inferred implicitly by the model.
Figure~\ref{fig:result4c} shows another example of our methods diversity and simultaneous optimization where the sphere allows for shape, texture and normal map optimization while the plane allows only for texture and normal map optimization. We show results for various distinct captions and also note that the texture and normal map of the plane optimize to support the object such as a picnic mat texture appearing when the caption is a "fruit basket on grass"

\subsection{Quantitative Evaluation}
We quantitatively evaluate our method, comparing it directly with the current closest work of ~\cite{jain2021zero}.
We follow the same experiment setup outlined in their paper: two shapes are generated per caption for a set of $153$ text captions, for a total of $306$ generated shapes. During evaluation they are rendered from a held-out pose not seen during training, a CLIP-R precision score \cite{Park2021BenchmarkFC} is then computed between the held-out pose renderings and the captions used to generate the shapes. The captions used are from \cite{jain2021zero} and the held out pose is also the same as theirs at a 45$^{\circ}$ elevation where as training is limited to a 30$^{\circ}$ elevations, we experiment with different sized CLIP models for generating the shapes and computing the precision.

Table \ref{tab:my-table} shows the quantitative results of the evaluation. 
Note that in this evaluation we do not include the diffusion prior loss as the dataset for training the prior contains only CLIP ViT-B/32 embeddings, so we are unable to train a prior that supports the generation model of CLIP ViT-B/16.
Regardless, we find that our work outperforms \cite{jain2021zero} across the generation and evaluation models without it.

\subsection{Ablation Studies}
In Table \ref{tab:ablated} an ablation study is shown for the various components of our pipeline. We start from a stripped down version of our method (baseline) and sequentially add in the components of the method. We follow the same evaluation methodology as in Table \ref{tab:my-table} but a single shape is generated per caption here instead of two as we found that it does not have a significant impact on the metric and reduces the time required per evaluation.
Our results show that the limit subdivision provides an improvement across all retrieval models. We then add the image augmentations which both provide improvements, offsetting the mesh from the center of the image provides the largest boost to the final results. Similarly, rendering the images at a higher resolution and then linearly scaling to the CLIP 224x224 resolution does improve results in all cases except for the largest ViT-L/14 model where it hurts performance. We get our best overall results when adding the prior loss.

%\begin{figure}
%\begin{center}
%\includegraphics[width=0.70\linewidth]{images/subdiv.png}
%\end{center}
%    \textcolor{red}{
%   \caption{Final output results (left column) with and (right column) %without limit subdivision for the text prompts "Burj Al Arab" (top) and "A %fruit basket" (bottom) }
%   }
%\label{fig:subdiv}
%\end{figure}

%\textcolor{red}{
%Our experiments also show that the limit subdivision improves the results both quantitatively and qualitatively. Table \ref{tab:ablated} shows the quantitative improvement and in Fig. \ref{fig:subdiv} we show meshes for the same text prompt with and without limit subdivision. Left is limit subdivision and right is without it - we find that it reduces mesh tangling and leads to better triangulation.
%}

% \usepackage{booktabs}
% \usepackage{multirow}

% \usepackage{booktabs}
% \usepackage{multirow}

% \begin{table}[]
% \begin{tabular}{@{}llc@{}}
% \toprule
% \multicolumn{2}{c}{\textbf{Method}} & \begin{tabular}[c]{@{}c@{}}R-Precision\\ CLIP ViT-B/32\end{tabular} \\ \midrule
% Baseline & COCO GT Images &  \\ \midrule
% Augmentations & + Background &  \\
% \multicolumn{1}{c}{} & + Reposition in Image &  \\ \midrule
% Render & + $512^2$ Renders & 65.3 \\ \midrule
% Prior & + Prior Loss & 78.4 \\
%  & + Select best prior &  \\ \bottomrule
% \end{tabular}
% \caption{Ablation evaluation}
% \label{tab:ablation}
% \end{table}
\begin{table}
    \centering
    \begin{tabular}{lcccc}
    % \begin{tabular}
    \hline 
    \textit{Generation Model} & \multicolumn{2}{c}{CLIP ViT-B/16}      & \multicolumn{2}{c}{CLIP ViT-B/32} \\ \hline
    \textit{Evaluation Model} &  ViT-B/16          &  ViT-B/32 &  ViT-B/16  &  ViT-B/32    \\ \hline
    Dreamfields        & 93.5 & 59.8 & 74.2   & 86.6    \\
    % Dreamfields        & 93.5 $\pm$ 1.4 & 59.8 $\pm$ 2.8 & 74.2   & 86.6    \\
     \cite{jain2021zero}        &  &  &    &    \\
    CLIP-Mesh [Ours]            & \textbf{96.7}        & \textbf{67.8} & \textbf{75.8}           &
    % Our Work                  & \textbf{96.7 $\pm$ 0.01}        & \textbf{67.8 $\pm$ 0.02} & \textbf{75.8}           &
    \textbf{91.4} \\ \hline
    \end{tabular}
    \caption{Quantative comparision of our work with dreamfields on COCO caption object generation}
    \label{tab:my-table}
\end{table}

\section{Conclusions, Limitations and Future Work}

We have demonstrated a method for generating diverse 3D objects in different modeling scenarios using only an input text prompt. 
The results consist of a mesh, texture map and normal map which allow them to be directly loaded to be used as assets in games and modelling applications. 
While the work we propose provides interesting results there are some limitations of our method.

\paragraph{Genus} The genus of the generated object is set by the initial template mesh. We address this issue partially by allowing a transparency channel in the texture, but a more principled approach is desirable.

\paragraph{CLIP Limitations} Using an image model to generate 3D shapes comes with its own challenges, since the model is trained with images it often projects artifacts to the mesh. Some examples of this can be seen in ~\ref{fig:result_wonders} where the pyramid has small people on its side and ~\ref{fig:result_comp2} where the mount Everest has the text "Everest" on its side and tip, note that we find using the larger CLIP ViT/B-32 model alleviates the text issue.

\begin{table}
\begin{tabular}{@{}lllll@{}}
\toprule
\multicolumn{2}{c}{{Method (CLIP B/16)}} & \multicolumn{3}{c}{CLIP R-Precision ↑}                            \\
\multicolumn{2}{c}{}                                    & \textit{ B/16} & \textit{ B/32} & \textit{ L/14} \\ \midrule
{Shape}           & Baseline Method      & 75.8               & 41.8               & 50.9               \\
                  & + Limit Subdivision  & 77.7               & 47.7               & 53.5               \\ \midrule
{Augmentations}   & + Background         & 81                 & 47.7               & 58.8               \\
                  & + Repositon Shape    & 90.1               & 60.5               & 73.2               \\ \midrule
Render            & + $512^2$ renders      & 92.1               & 62.7               & 70.5               \\ \midrule
Prior             & + Prior Loss         & 91.5               & 77.7               & 74.5               \\ \bottomrule
\end{tabular}
\caption{Ablation study on the R-Precision quantitative metric where higher score is better. We observe that starting from a baseline approach, adding limit subdivision, augmentation, large rendering, and the generative prior systematically improves performance}
\label{tab:ablated}
\end{table}
In future work we will aim to further improve shape based constraints and explore methods to provide more user control in the generative process.

%\Acknowledgment

\section{Acknowledgements}

We acknowledge the support of the Natural Sciences and Engineering Research Council of Canada (NSERC) [RGPIN-2021-04104 and RGPIN-2021-03477]

This research was enabled in part by support provided by Calcul Quebec (calculquebec.ca) and the Digital Research Alliance of Canada (alliancecan.ca).

 \newpage
\bibliographystyle{ACM-Reference-Format}
\bibliography{main}

\end{document}